\documentclass{ws-procs11x85}
\usepackage{lmodern}

\usepackage[T1]{fontenc}
\usepackage[latin9]{inputenc}
\usepackage{amstext}
\usepackage{graphicx}

\makeatletter

\providecommand{\tabularnewline}{\\}

\makeatother

\begin{document}
\title{EVALUATION OF LINEAR CLASSIFIERS ON ARTICLES CONTAINING PHARMACOKINETIC EVIDENCE  OF DRUG-DRUG INTERACTIONS}

\author{A. KOLCHINSKY}
\address{School of Informatics and Computing, Indiana University\\ Bloomington, IN, USA\\ E-mail: akolchin@indiana.edu}

\author{A. LOURENÇO}
\address{Institute for Biotechnology \& Bioengineering, Centre of Biological Engineering, University of Minho\\Braga, Portugal\\E-mail:analia@deb.uminho.pt}

\author{L. LI }
\address{Department of Medical and Molecular Genetics, Indiana Univeristy School of Medicine\\ Indianapolis, IN, USA\\ E-mail: lali@iupui.edu}

\author{L. M. ROCHA$^*$ }
\address{School of Informatics and Computing, Indiana University\\ Bloomington, IN, USA\\ E-mail: rocha@indiana.edu}
\begin{abstract}
\textbf{Background}. Drug-drug interaction (DDI) is a major cause
of morbidity and mortality. DDI research includes the study of different
aspects of drug interactions, from \emph{in vitro} pharmacology, which
deals with drug interaction mechanisms, to pharmaco-epidemiology,
which investigates the effects of DDI on drug efficacy and adverse
drug reactions. Biomedical literature mining can aid both kinds of
approaches by extracting relevant DDI signals from either the published
literature or large clinical databases. However, though drug interaction
is an ideal area for translational research, the inclusion of literature
mining methodologies in DDI workflows is still very preliminary. One
area that can benefit from literature mining is the automatic identification
of a large number of potential DDIs, whose pharmacological mechanisms
and clinical significance can then be studied via \emph{in vitro}
pharmacology and \emph{in populo }pharmaco-epidemiology. 

\textbf{Experiments.} We implemented a set of classifiers for identifying
published articles relevant to experimental pharmacokinetic DDI evidence.
These documents are important for identifying causal mechanisms behind
putative drug-drug interactions, an important step in the extraction
of large numbers of potential DDIs. We evaluate performance of several
linear classifiers on PubMed abstracts, under different feature transformation
and dimensionality reduction methods. In addition, we investigate
the performance benefits of including various publicly-available named
entity recognition features, as well as a set of internally-developed
pharmacokinetic dictionaries.

\textbf{Results.} We found that several classifiers performed well
in distinguishing relevant and irrelevant abstracts.  We found that
the combination of unigram and bigram textual features gave better
performance than unigram features alone, and also that normalization
transforms that adjusted for feature frequency and document length
improved classification. For some classifiers, such as linear discriminant
analysis (LDA), proper dimensionality reduction had a large impact
on performance. Finally, the inclusion of NER features and dictionaries
was found not to help classification. 
\end{abstract}

\section{Introduction}

Drug-drug interaction (DDI) has been implicated in nearly 3\% of all
hospital admissions \cite{jankel1993epidemiology} and 4.8\% of admissions
among the elderly \cite{becker2007hospitalisations}; it is also a
common form of medical error, representing 3\% to 5\% of all inpatient
medication errors \cite{leape1995systems}. With increasing rates
of polypharmacy, which refers to the use of multiple medications or
more medications than are clinically indicated \cite{hajjar2007polypharmacy},
the incidence of DDI will likely increase in the coming years.

DDI research includes the study of different aspects of drug interactions.
\emph{In vitro} pharmacology experiments use intact cells (e.g. hepatocytes),
microsomal protein fractions, or recombinant systems to investigate
drug interaction mechanisms. Pharmaco-epidemiology (\emph{in populo})
uses a population based approach and large electronic medical record
databases to investigate the contribution of a DDI to drug efficacy
and adverse drug reactions. 

Biomedical literature mining (BLM) can be used to detect novel DDI
signals from either the published literature or large clinical databases
\cite{tatonetti2011detecting}. BLM is becoming an important biomedical
informatics methodology for large scale information extraction from
repositories of textual documents, as well as for integrating information
available in various domain-specific databases and ontologies, ultimately
leading to knowledge discovery \cite{shatkay2003mining,jensen2006literature,cohen2008getting}.
It has seen applications in research areas that range from protein-protein
interaction \cite{leitner2010febs,krallinger2011protein}, protein
structure \cite{rechtsteiner2010use}, genomic locations associated
with cancer \cite{mcdonald2004entity}, drug targets \cite{el2008mining},
and many others. BLM holds the promise of tapping into the biomedical
collective knowledge and uncovering relationships buried in the literature
and databases, especially those relationships present in global information
but unreported in individual experiments \cite{abi2008uncovering}. 

Although pharmaco-epidemiology and BLM approaches are complementary,
they are usually conducted independently. DDI is thus an exemplary
case of translational research that can benefit from interdisciplinary
collaboration. In particular, automated literature mining methods
allow for the extraction of a large number of potential DDIs whose
pharmacological mechanisms and clinical significance can be studied
in conjunction with \emph{in vitro }pharmacology and \emph{in populo
}pharmaco-epidemiology.

Though BLM has previously been used for DDI information extraction
\cite{segura2010resolving,percha2012discovery}, much remains to be
done before it can integrated into translational workflows. One gap
is in the extraction of DDI information from a pharmacokinetics perspective,
since existing methods do not explicitly capture pharmacokinetics
parameters and do not consider knowledge from \emph{in vitro }and
\emph{in vivo} DDI experimental designs, especially the selection
of enzyme-specific probe substrates and inhibitors. For instance,
important pharmacokinetic parameters such as Ki, IC50, and AUCR have
not been included in existing text mining approaches to DDI. Yet this
kind of pharmacokinetic information may be particularly relevant when
seeking evidence of causal mechanisms behind DDIs, and as a complement
to DDI text mining of patient records, where reporting biases and
confounds often give rise to non-causal correlations \cite{tatonetti2012data}. 

We have previously showed that BLM can be used for automatic extraction
of numerical pharmacokinetics (PK) parameters from the literature
\cite{wang2009literature}. However, that work was not oriented specifically
toward extraction of DDI information. In order to perform DDI information
extraction from a pharmacokinetics perspective, we first need to be
able to identify the relevant documents that contain such information.
Here, we evaluate the performance of text classification methods on
documents that may contain pharmacology experiments in which evidence
for DDIs is reported. Our goal is to develop and evaluate automated
methods of identifying DDIs backed by reported pharmacokinetic evidence,
which we believe is an essential first step towards the integration
of literature mining methods into translational DDI workflows. A collaboration
between Rocha's lab, working on BLM, and Li's lab, working on \emph{in
vitro }pharmacokinetics, was developed in order to pursue this goal.

In this paper, we report on the performance of a set of classifiers
on a manually-annotated  corpus produced by Li's lab. We consider
a wide range of linear classifiers, among them logistic regression,
support vector machines (SVM), binomial Naive Bayes, linear discriminant
analysis, and a modification of our `Variable Trigonometric Threshold'
(VTT) classifier, which was previously found to perform well on protein-protein
interaction text mining tasks \cite{lourencco2011linear,kolchinsky2010classification,abi2008uncovering}.
In addition, we compare different feature transformation methods,
including normalization techniques such as TFIDF and PCA-based dimensionality
reduction. We also compare performance when using features generated
by several Named Entity Recognition (NER) tools.

In the next section, we describe the  corpus used in this study.
Section \ref{sec:Classifiers} discusses the evaluated classifiers,
while section \ref{sec:Feature-Transforms} deals with dimensionality
reduction and feature transforms. Section \ref{sec:Performance-evaluation}
covers our methods of cross-validation and performance evaluation.
Section \ref{sec:Results} provides classification performance results
for textual features, while section \ref{sec:NER-performance} does
so for the combination of textual and NER features. We conclude with
a discussion in section \ref{sec:Discussion}.

\section{Corpus\label{sec:Corpus}}

Li's lab selected 1213 PubMed pharmacokinetics-related abstracts for
the training corpus. Documents were obtained by first searching PubMed
using terms from an ontology previously developed for automatic extraction
of numerical PK pharmacokinetics parameters \cite{wang2009literature}.
The retrieved articles were manually classified into two groups: abstracts
that explicitly mentioned evidence for the presence or absence of
drug-drug interactions were labeled as DDI-relevant (602 abstracts),
while the rest were labeled as DDI-irrelevant (611 abstracts). DDI-relevance
was established if articles contained one of the four primary classes
of pharmacokinetics studies: clinical PK studies, clinical pharmacogenetic
studies, \textit{in vivo} DDI studies, and \textit{in vitro} drug
interaction studies. The classification was initially done by three
graduate students with M.S. degrees and one postdoctoral annotator.
Any inter-annotator conflicts were further checked by a Pharm D. and
an M.D. scientist with extensive pharmacological training. The corpus,
as well as further details \cite{wu2012integrated}, is available
upon request.

We extracted textual features from the abstract title and abstract
text, as well as several other PubMed fields. These included the author
names, the journal title, the Medical Subject Heading (MeSH) terms,
the \textquoteleft{}registry number/EC number\textquoteright{} (RN)
field, and the \textquoteleft{}secondary source\textquoteright{} field
(SI) (the latter two contain identification codes for relevant chemical
and biological entities). For each PubMed entry, the content of the
above fields was tokenized, processed by Porter stemming, and converted
into textual features (unigrams and, in certain runs, bigrams). Strings
of numbers were converted into `\#', while short textual features
(those with a length of less than 2 characters) and infrequent features
(those that occurred in less than 2 documents) were omitted. Each
MeSH term was treated as a single textual token. Finally, the occurrence
of different features in different documents was recorded in binary
occurrence matrices. We evaluated performance using unigram features
only (the unigram runs), as well as using a combination of unigram
and bigram features (the bigram runs).

\section{Classifiers\label{sec:Classifiers}}

Six different linear classifiers were implemented:
\begin{enumerate}
\item VTT: a simplified, angle-domain version of our \emph{`}Variable Trigonometric
Threshold' Classifier (VTT) \cite{lourencco2011linear,kolchinsky2010classification,abi2008uncovering}.
Given a binary document vector $\mathbf{x}=\left\langle x_{1},\dots,x_{K}\right\rangle $,
with its features (i.e. dimensions) indexed by $i$, the VTT separating
hyperplane is:
\[
\sum_{i}\theta_{i}x_{i}-\lambda=0
\]
Here, $\lambda$ is a threshold (bias) and $\theta_{i}$ is the `angle'
of feature $i$ in class space:
\[
\theta_{i}=\arctan\frac{p_{i}}{n_{i}}-\frac{\pi}{4}
\]
where $p_{i}$ is the proportion of positive-class documents in which
feature $i$ occurs, and $n_{i}$ is the proportion of negative-class
documents in which features $i$ occurs. $\theta_{i}$ is positive
when $p_{i}\ge n_{i}$ and negative otherwise. The threshold parameter
$\lambda$ is chosen via cross-validation. The full version of VTT,
previously used in protein-protein interaction tasks, includes additional
parameters to account for named entity occurrences and is used in
section \ref{sec:NER-performance} below. VTT performs best on sparse
data sets, in which most feature values $x_{i}$ are set to 0; for
this reason, we do not evaluate it on dense dimensionality-reduced
datasets (see below).
\item SVM: a linear Support Vector Machine (SVM) classifier (provided by
the \textsf{sklearn} \cite{scikit-learn} library's interface to the
\textsf{LIBLINEAR} package \cite{fan2008liblinear}) with a cross-validated
regularization parameter.
\item Logistic regression: a logistic regression classifier (also provided
by \textsf{sklearn}'s interface to \textsf{LIBLINEAR}) with a cross-validated
regularization parameter. 
\item Naive Bayes: a binomial Naive Bayes classifier with a Beta-distributed
prior for smoothing. The prior's concentration parameter was determined
by cross-validation.
\item LDA: a Linear Discriminant Analysis (LDA) classifier, where the data
covariance matrix was shrunk toward a diagonal, equal-variance structured
estimate. The shrinkage parameter was determined by cross-validation.
\item dLDA: a \textquoteleft{}diagonal\textquoteright{} version of LDA,
where only the diagonal entries of the covariance matrix are estimated
and the off-diagonal entries are taken to be 0. A cross-validated
parameter determines shrinkage toward a diagonal, equal-variance estimate.
This classifier provides a more robust estimate of feature variances;
it is equivalent to a Naive Bayes classifier for multivariate Gaussian
features \cite{bickel2004some}.
\end{enumerate}

\section{Feature Transforms\label{sec:Feature-Transforms}}

For both unigram and bigram runs, the classifiers were applied to
the following data matrices:
\begin{enumerate}
\item No transform: the raw binary occurrence matrices, as described in
section \ref{sec:Corpus}. For LDA, when the number of documents ($N$)
was less than the number of dimensions (giving rise to singular covariance
matrices), the occurrence matrices were projected onto their first
$N$ principal components.
\item IDF: occurrences of feature $i$ were transformed into that feature's
Inverse Document Frequency (IDF) value: 
\[
\textrm{idf}\left(i\right)=\log\frac{N}{c_{i}+1}
\]
where $c_{i}$ is the total number of occurrences of features $i$
among all documents. This reduced the influence of common words on
classification.
\item TFIDF: the Term Frequency, Inverse Document Frequency (TFIDF) transform
applies the above IDF transform, and then divides each document's
feature values by the total number of that document's features. This
attempts to minimize differences between documents of different sizes
(i.e. with different numbers of features).
\item Normalization: here the non-transformed, IDF, and TFIDF document matrices
underwent a length-normalization transform, where each document vector
was inversely scaled by its L2 norm. This normalization has been argued
to be especially important for good SVM performance \cite{leopold2002text}.
\item PCA-based dimensionality reduction: The above matrices were run through
a Principal Component Analysis (PCA) dimensionality reduction step.
Projections onto the first 100, 200, 400, 600, 800, and 1000 components
were applied.
\end{enumerate}

\section{Performance evaluation\label{sec:Performance-evaluation}}

We evaluated the performance of the classifiers using three different
measures: the commonly-used F1 score, the area under the interpolated
precision/recall curve \cite{davis2006relationship} (here called
iAUC), and Matthews Correlation Coefficient \cite{matthews1975comparison}
(MCC).

In this task, only one corpus was provided. Thus, we had to use it
both for training classifiers and for measuring generalization performance
on out-of-sample documents. We performed the following cross-validation
procedure to estimate generalization performance:
\begin{enumerate}
\item The documents of the entire corpus were partitioned into 4 folds (75\%-25\%
splits). This was repeated 4 times, giving a total of 16 folds (we
call these the \emph{outer folds}).
\item For each fold, classifiers were trained on 75\% block of the corpus
and tested on the 25\% block of the corpus.
\item The 16 sets of testing results were averaged to produce an estimate
of generalization performance.
\end{enumerate}
In addition, all of the classifiers mentioned in section \ref{sec:Classifiers}
contain cross-validated parameters: for VTT, this is the bias parameter,
while the other classifiers have regularization or smoothing parameters.
In order to fully separate training from testing data and accurately
estimate generalization performance, nested cross-validation was done
within each of the 75\% blocks of the above outer folds:
\begin{enumerate}
\item The 75\% block is itself partitioned into 4 folds (75\%-25\% splits
of the 75\% block). This is repeated 4 times, producing a total of
16 folds (we call these the \emph{inner folds})
\item For each searched value of the cross-validated parameter, a classifier
is trained on each of the 16 inner folds' 75\% block and tested on
its 25\% block. 
\item The value giving the best average performance (here, according to
the MCC metric) is chosen as the cross-validated parameter value for
this outer fold.
\end{enumerate}
An outer fold's cross-validated parameter value is then used to train
on the fold's 75\% block and test on its 25\% block.

\section{Classification performance \label{sec:Results}}

\subsection{Overall performance}

\begin{figure}
\hfill{}

\begin{minipage}[t]{0.5\columnwidth}%
\includegraphics[width=1\textwidth]{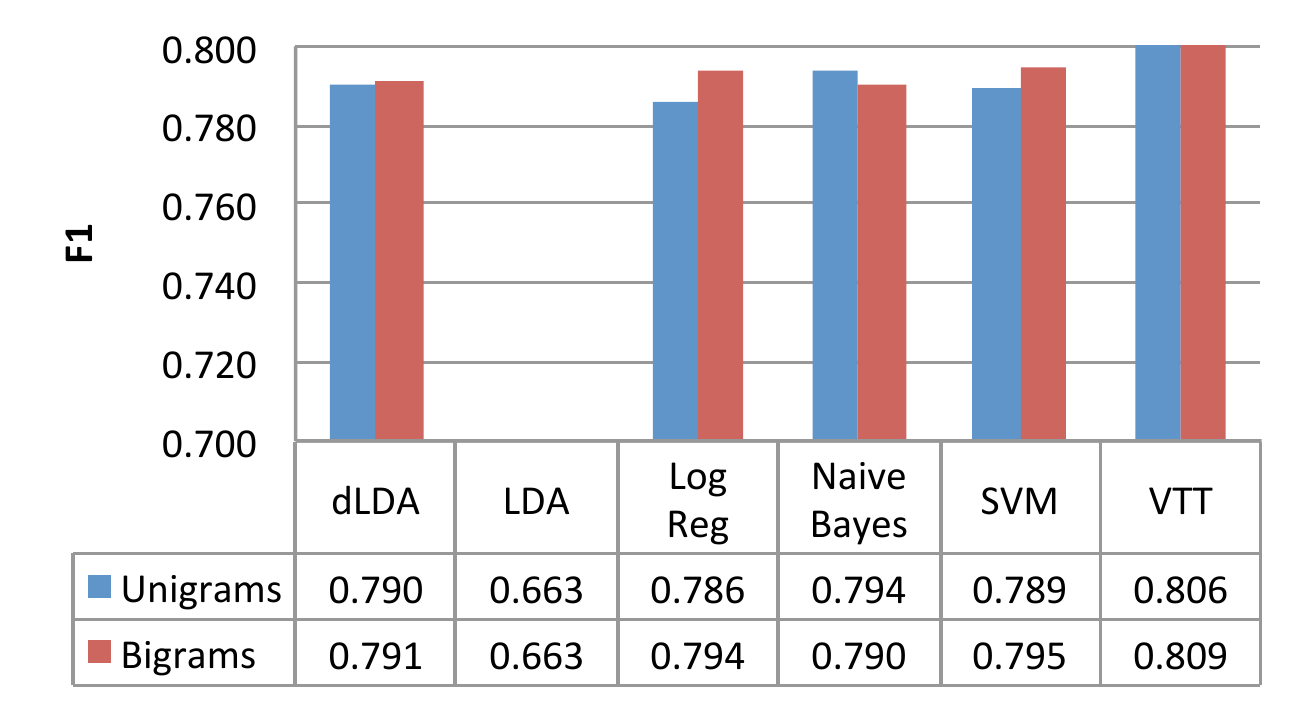}%
\end{minipage}%
\begin{minipage}[t]{0.5\columnwidth}%
\includegraphics[width=1\textwidth]{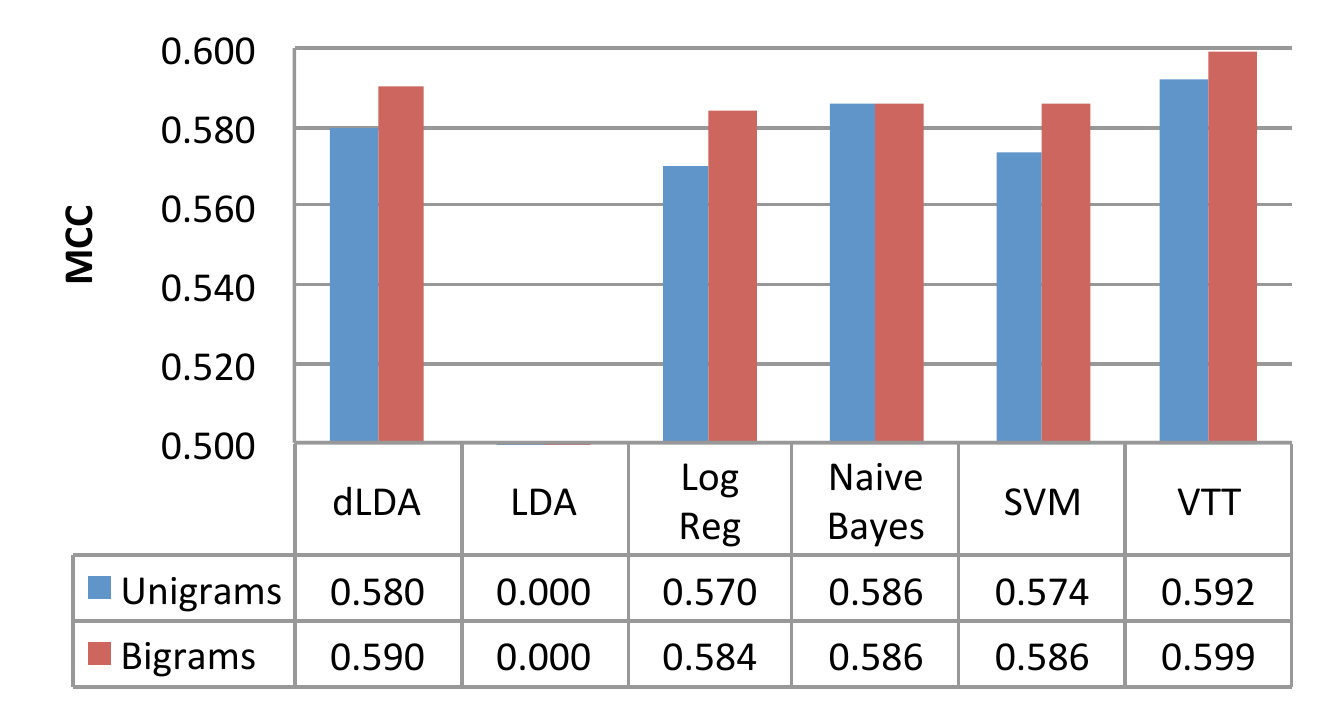}%
\end{minipage}

\begin{minipage}[t][1\totalheight][c]{0.5\columnwidth}%
\includegraphics[width=1\textwidth]{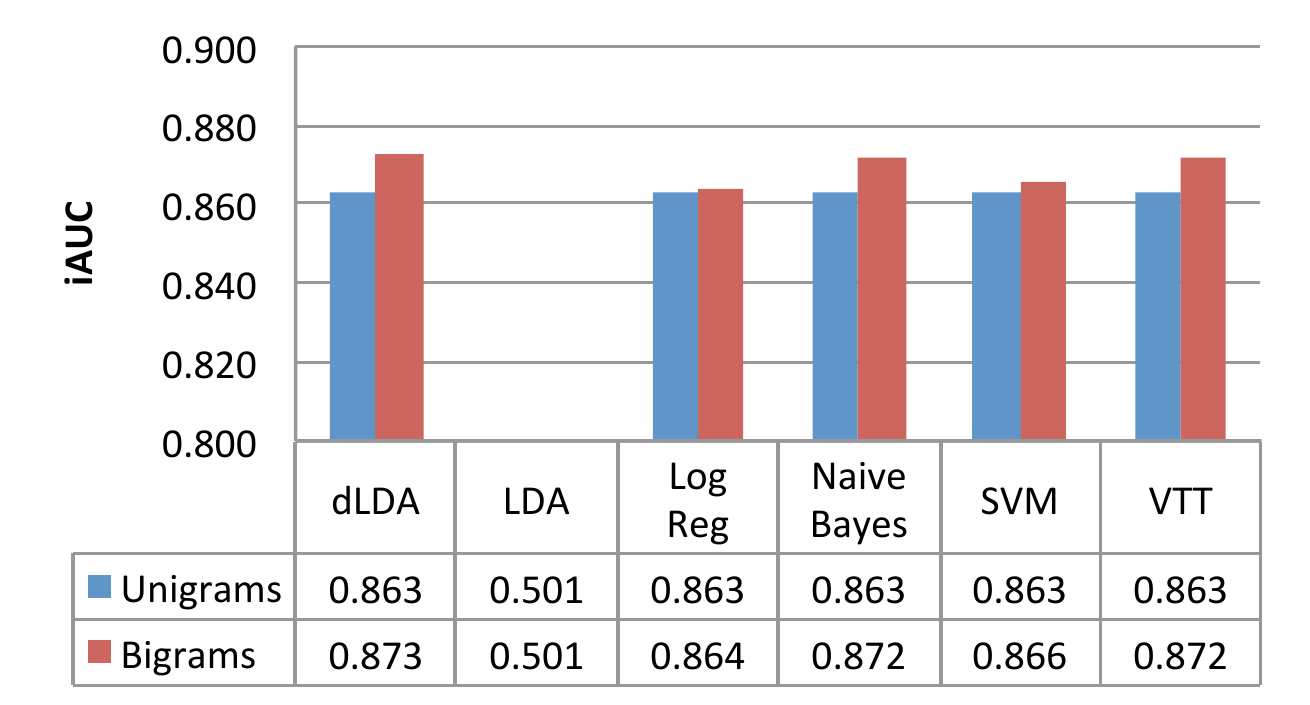}%
\end{minipage}%
\begin{minipage}[t][1\totalheight][c]{0.5\columnwidth}%
\begin{center}
\begin{tabular}{|c|}
\hline 
\texttt{\scriptsize dai}\tabularnewline
\texttt{\scriptsize \# mg}\tabularnewline
\texttt{\scriptsize mg}\tabularnewline
\emph{\scriptsize MeSH: Cross-Over Studies}\tabularnewline
\texttt{\scriptsize on dai}\tabularnewline
\emph{\scriptsize MeSH: Drug Interactions}\tabularnewline
\texttt{\scriptsize crossov studi}\tabularnewline
\texttt{\scriptsize crossov}\tabularnewline
\texttt{\scriptsize random}\tabularnewline
\texttt{\scriptsize daili}\tabularnewline
\hline 
\end{tabular}
\par\end{center}%
\end{minipage}\caption{Classification performance using non-transformed features, for both
unigram and bigram runs. Top left is the F1 measure, top right is
the MCC measure, and lower left is the iAUC measure. LDA performed
poorly and is below the charts' lower cutoff value. Lower right shows
the top 10 features identified in a typical bigram fold, ranked according
to the information gain criteria.\label{fig:performance}}
\end{figure}

Figure \ref{fig:performance} shows the performance of the classifiers
in unigram runs (which included only unigram features) and bigram
runs (which included both unigram and bigram features), without any
feature transforms applied. In addition, it also shows the top 10
features identified in a typical bigram fold, ranked according to
the information gain criteria \cite{yang1997comparative}.

With the exception of LDA, all of the classifiers performed similarly
on the task. VTT performed slightly better than the other classifiers
according to the F1 and MCC measures. LDA's performance was dismal,
suggesting that in such a high-dimensional setting there is not enough
data to estimate the feature covariance matrix, even under covariance
matrix shrinkage. This is supported by the fact that the dLDA (diagonal
LDA) classifier, which estimates only the diagonal entries of the
covariance matrix, performed well on the task.

The difference between unigram and bigram runs was not major, but
bigram performance showed a consistent small improvement, indicating
that the advantage in predictability provided by bigrams outweighs
their cost in additional parameters. For the rest of this work, we
will only report on the bigram run performance. The pattern of performance
for the unigram runs was similar to that of bigram runs.

\subsection{Feature transforms}

\begin{figure}
\begin{minipage}[t]{0.55\columnwidth}%
\includegraphics[bb=30bp 0bp 590bp 255bp,width=1\columnwidth]{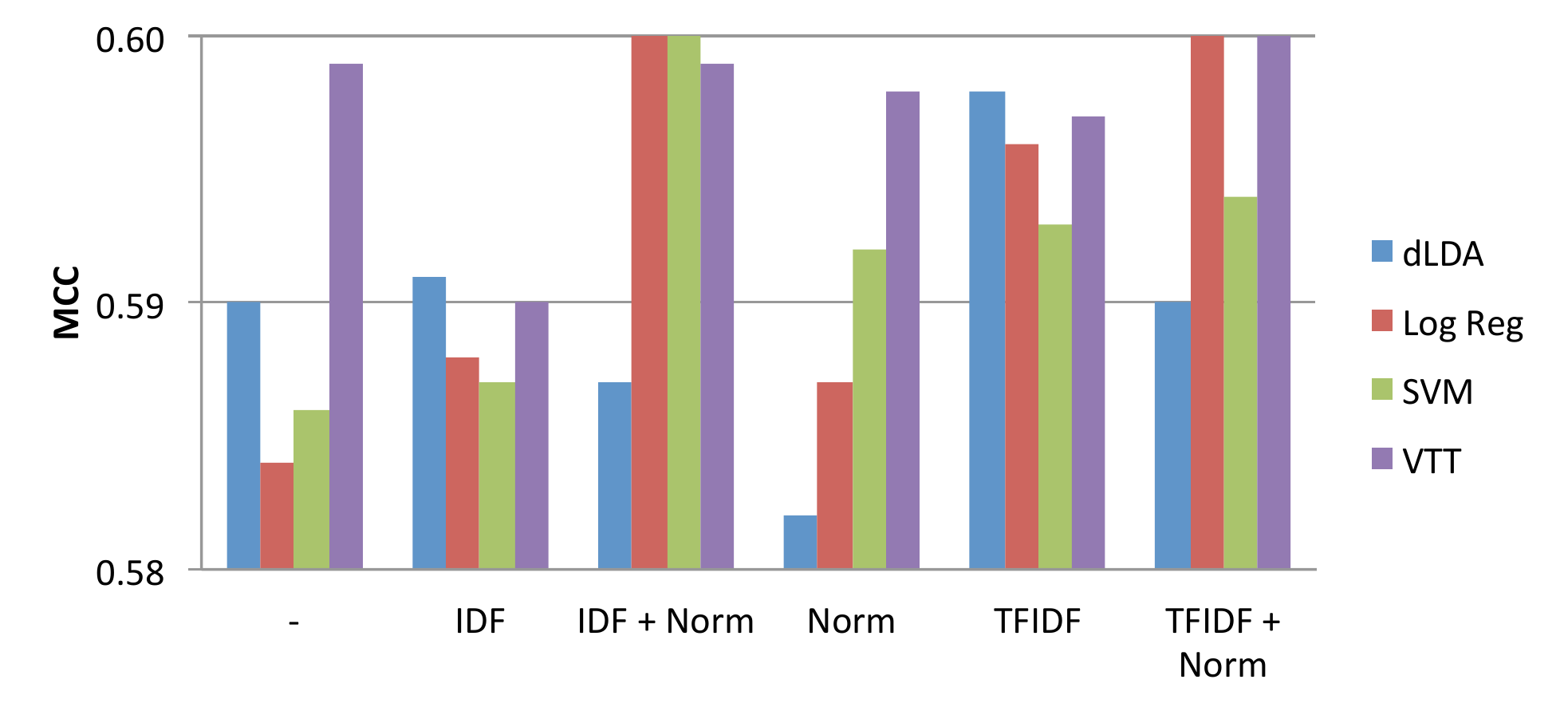}%
\end{minipage}%
\begin{minipage}[t]{0.39\columnwidth}%
\includegraphics[bb=30bp 0bp 380bp 203bp,width=1\columnwidth]{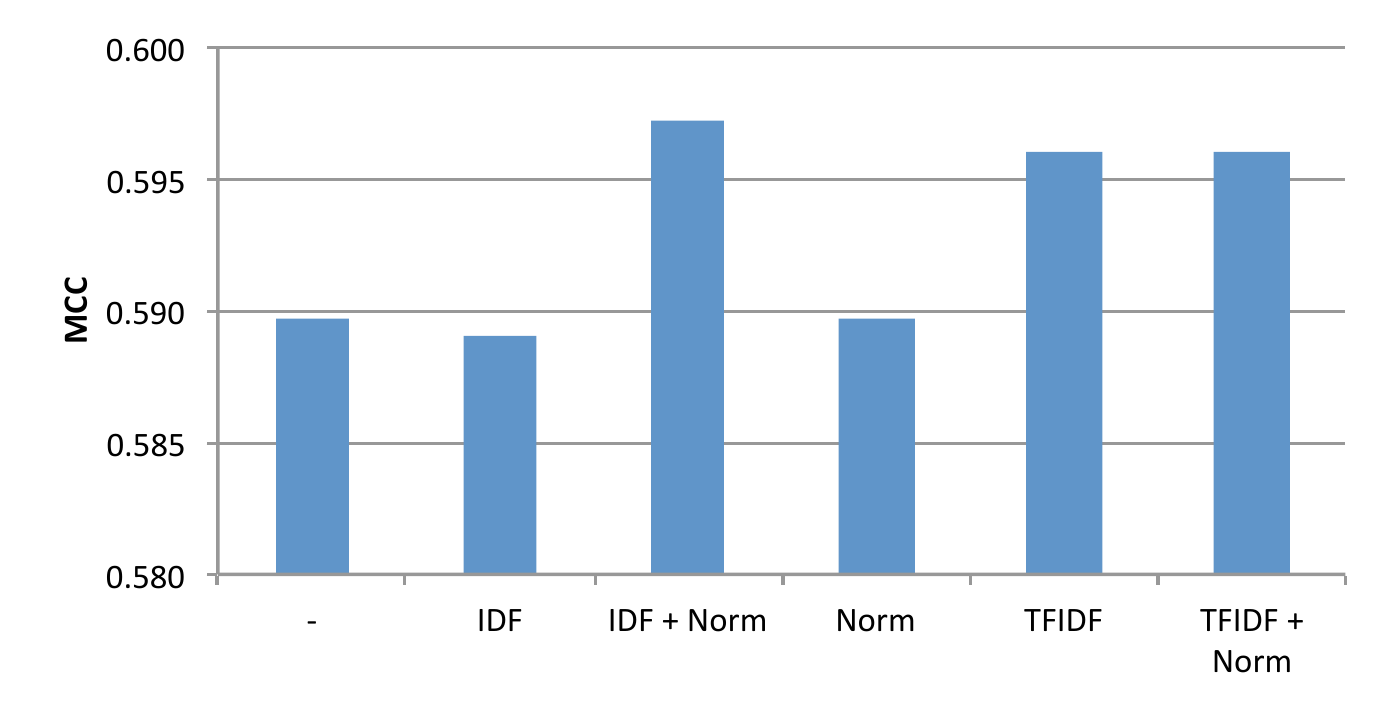}%
\end{minipage}

\caption{MCC performance using bigram features under various transforms. `-'
refers to no transform, IDF and TFIDF refer to transforms described
in section \ref{sec:Feature-Transforms}, while IDF+Norm and TFIDF+Norm
refer to those same transforms followed by unit-length normalizations.
Results are shown for 4 well-performing classifiers (left); average
MCC values across those 4 classifiers (right).\label{fig:features}}
\end{figure}

For simplicity, in the following sections we present performance results
in terms of MCC values only. It is important to note that in most
of the conditions, the 16-fold estimate of MCC performance gave a
standard error on the order of 0.01; differences in performance of
this scale can be ascribed to statistical fluctuations.

In figure \ref{fig:features}, we plot the performance of the classifiers
under different feature transform methods on the bigram runs. We tested
these transforms under 4 classifiers: diagonal LDA (dLDA), SVM, Logistic
Regression (Log Reg), and VTT. LDA performance is not reported, since
as previously seen it performs badly on high-dimensional data. The
binomial Naive Bayes classifier was omitted because it is not applicable
to non-binary data.

The different transforms did not change performance dramatically,
but some did offer advantages. VTT performed consistently well across
different kinds of transforms, except for the IDF transform, where
its performance decreased. As expected, SVM benefited from length
normalization (whether L2-type unit-length normalization, or L1-type
normalization offered by the term-frequency part of TFIDF). As seen
in the bottom section of figure \ref{fig:features}, the transforms
offering good performance across a range of classifiers seemed to
be those combining an IDF correction with some kind of length normalization:
either IDF+Norm or TFIDF (with or without unit-length normalization).

\subsection{Dimensionality reduction}

\begin{figure}[t]
\begin{centering}
\includegraphics[height=8cm]{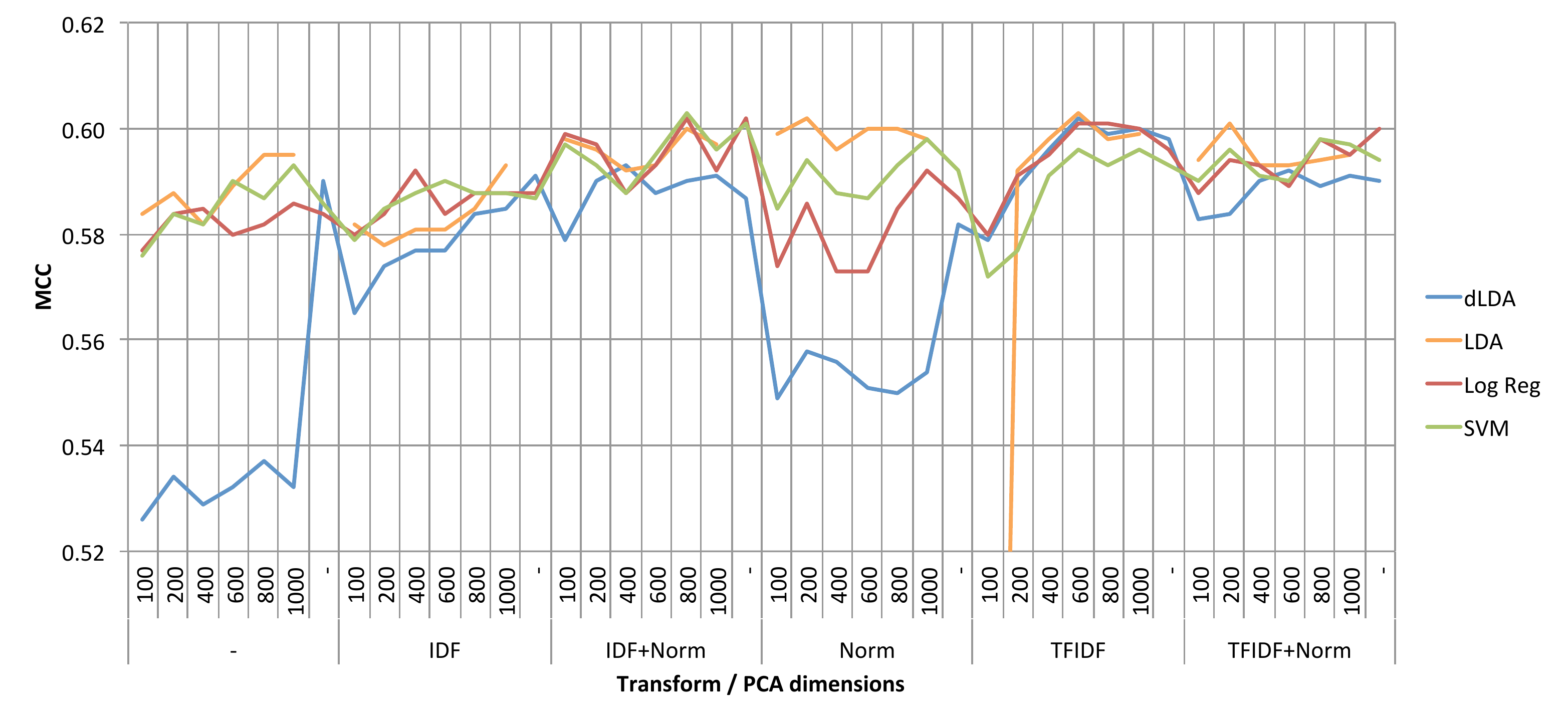}
\par\end{centering}

\caption{\label{fig:pca-perf}MCC performance on abstracts under different
feature transforms and PCA-based dimensionality reductions, bigram
runs. The very bottom lists different transforms, while the numbers
refer to the number of principal components kept. `-' refers to
both no transform (original data matrix) and to no dimensionality
reduction, as appropriate.}
\end{figure}

Figure \ref{fig:pca-perf} shows the performance of 4 classifiers
under PCA-based dimensionality reduction on the bigram runs. Here,
after applying the previously described transforms, the data matrices
are projected onto their principal components. This generates smaller-dimensional,
non-sparse data matrices. In this case, we have omitted the VTT classifier,
since it does not generalize to non-sparse datasets. We have also
omitted the binomial Naive Bayes classifier, since it is not applicable
to non-binary data.

\begin{figure}[t]
\begin{centering}
\includegraphics[height=6cm]{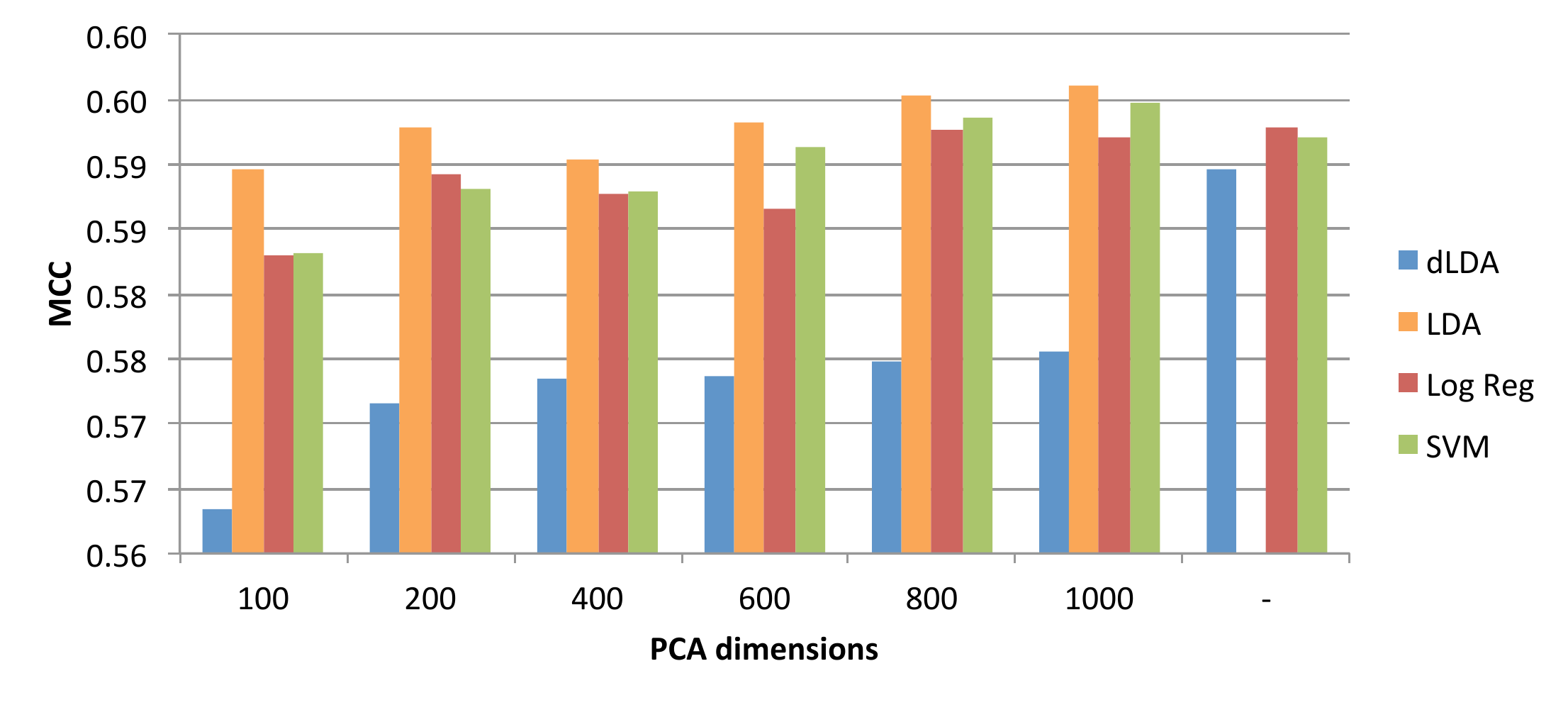}
\par\end{centering}

\caption{\label{fig:pca-perf-avg-by-classifier}MCC performance of different
classifiers under feature transforms and dimensionality reduction
condition, but now averaged across different feature transforms, bigram
runs. The bottom axis refers to number of principal components kept,
and `-' refers to no dimensionality reduction.}
\end{figure}

Dimensionality reduction only has a significant effects on performance
for LDA, where this is expected. Because LDA requires an estimate
of the full feature covariance matrix, it does not perform well when
the data is very high-dimensional (and hence, the covariance matrix
is difficult to estimate). However, under dimensionality reduction
LDA performs extremely well, often outperforming other classifiers.
Figure \ref{fig:pca-perf-avg-by-classifier} shows the performance
of different classifiers under different dimensionality reductions,
now averaged across the 6 feature transforms described previously.
Interestingly performance tends to increase as more principal components
are kept. With 1000 principal components, LDA has the best on-average
performance, though SVM also does well here. On the other hand, Diagonal
LDA -- which does not take into account feature covariances -- does
not perform well under dimensionality reduction.

\section{Classification performance on abstracts with NER\label{sec:NER-performance}}

The above runs used the occurrences of unigrams and bigrams as features.
We have previously used features extracted using Named Entity Recognition
(NER) tools in order to improve classification performance on a protein-protein
interaction text mining task \cite{lourencco2011linear,kolchinsky2010classification,abi2008uncovering}.
NER identifies occurrences of named entities (for example, drugs,
proteins, or chemical names) in documents. We applied a set of NER
extraction tools and used the count of named entities identified in
each document as an additional document feature, on top of the textual
occurrence features previously discussed.

The following publicly-available tools were used to identify named
entities:
\begin{itemize}
\item OSCAR4 \cite{jessop2011oscar4}: a recognizer of chemical names
\item ABNER \cite{settles2005abner}: biomedical named entity recognizer
for proteins
\item DrugBank \cite{wishart2006drugbank}: a database of drug names
\item BICEPP \cite{lin2011bicepp}: a recognizer of clinical characteristics
associated with drugs 
\end{itemize}
We also identified named entities using the following dictionaries,
provided by Li's lab \cite{wu2012integrated}:
\begin{itemize}
\item i-CYPS: a dictionary of cytochrome P450 {[}CYP{]} protein names, a
group of enzymes centrally involved in drug metabolism
\item i-PkParams: a dictionary of pharmacokinetic parameters
\item i-Transporters: a dictionary of proteins involved in transport
\item i-Drugs: a dictionary of Food and Drug Administration's drug names
\end{itemize}
For SVM, Logistic Regression, and LDA, the NER counts were treated
as any other feature. Diagonal LDA was omitted since it was outperformed
by dimensionality-reduced LDA, and binomial Naive Bayes was omitted
since NER-count features are non-binary. VTT incorporates NER-count
features via a modified separating hyperplane equation:
\[
\sum_{i}\theta_{i}x_{i}-\sum_{j}\frac{\beta_{j}-c_{j}}{\beta_{j}}-\lambda=0
\]
where $x_{i}$ represent non-NER feature occurrences, $\theta_{i}$
and $\lambda$ are textual feature weighting and bias parameters as
described in section \ref{sec:Classifiers}, $c_{j}$ is the count
of NER features produced for the current document by NER tool $j$,
and $\beta_{j}$ is a cross-validated weighting term for NER tool
$j$.

The classifiers were run on occurrence matrices with no transform
applied, except for LDA, which was run on occurrence matrices projected
onto their first 800 principal components. It is important to note
that in the presence of NER count features, whose values are of a
different magnitude from those of binary occurrence features, length
normalization can significantly hurt classifier performance (data
not shown).

\begin{figure}[t]
\begin{centering}
\includegraphics[width=0.8\textwidth]{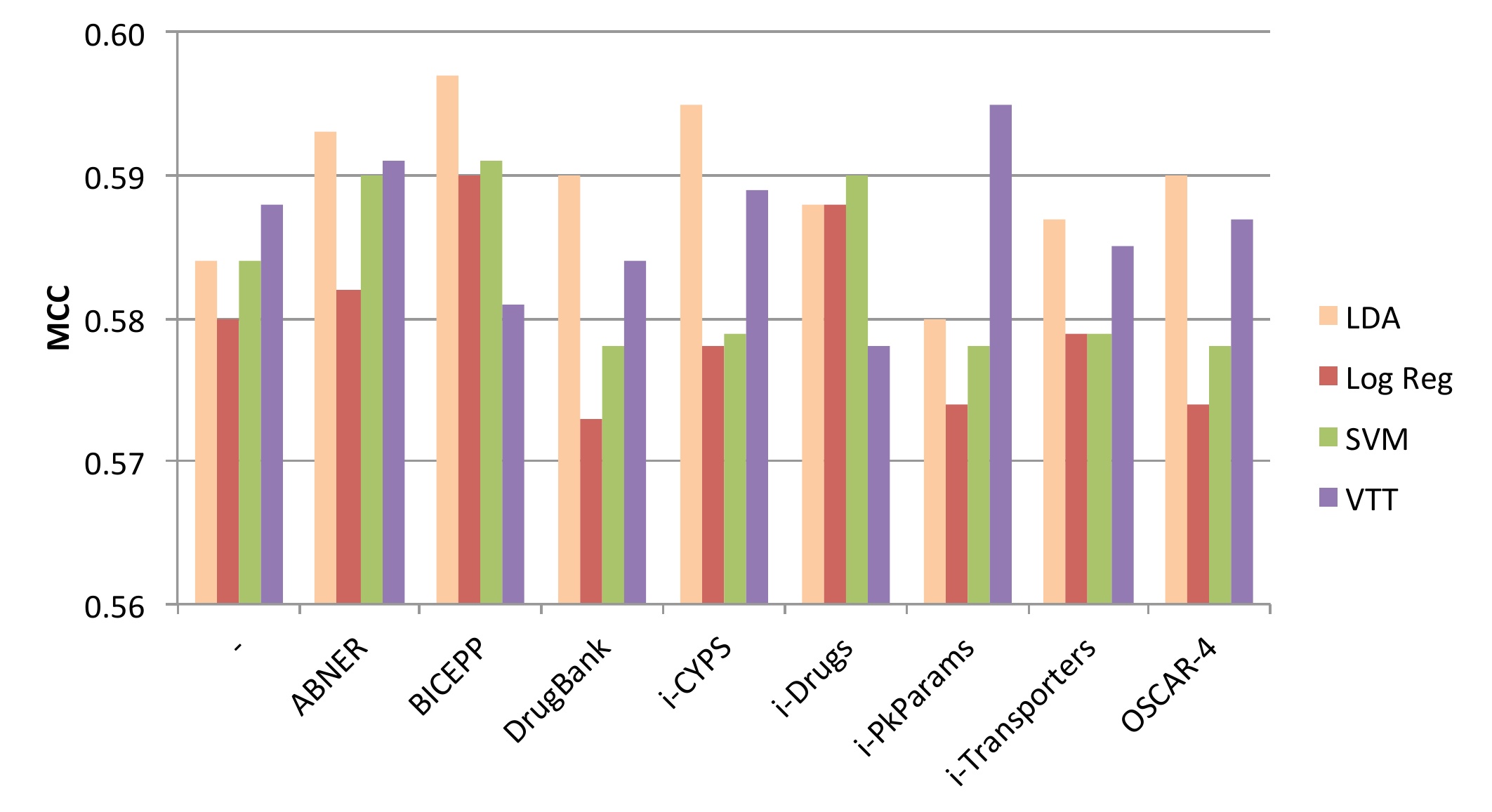}
\par\end{centering}

\caption{\label{fig:NER-performance}MCC performance of the classifiers in
combination with different NER features on the bigram runs. Classifiers
used non-transformed data matrices, apart from LDA which was applied
to an occurrence matrix projected onto its first 800 principal components.}
\end{figure}

\begin{figure}[t]
\begin{centering}
\includegraphics[height=5cm]{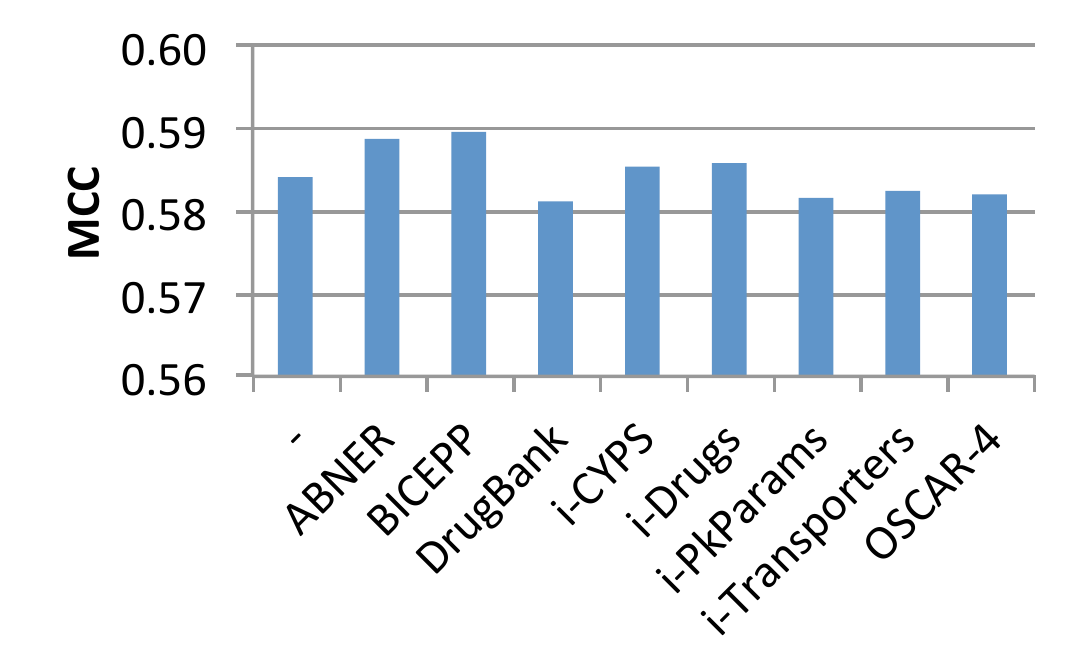}
\par\end{centering}

\caption{\label{fig:NER-performance-avg}MCC performance when using NER features
on the bigram runs, averaged across the 4 classifiers shown in figure
\ref{fig:NER-performance}.}
\end{figure}

Figure \ref{fig:NER-performance} shows the performance of the different
classifiers on a combination of bigram and NER features, while figure
\ref{fig:NER-performance-avg} shows the same performance averaged
across classifiers. Given the scale of standard errors of MCC performance
estimates (\textasciitilde{}0.01), it does not appear that NER features
offer a significant improvement in classification rates. We also attempted
to use combinations (pairs) of NER features in classification, but
this also failed to improve performance (data not shown). We discuss
possible reasons for this in the final section.

\section{Discussion\label{sec:Discussion}}

We studied the performance of BLM on the problem of automatically
identifying DDI-relevant PubMed abstracts, that is those containing
pharmacokinetic evidence for the presence or absence of drug-drug
interactions (DDI). We compared the performance of several linear
classifiers  using different combinations of unigrams, bigrams, and
NER features. We also tested the effect several feature transformation
and normalization methods, as well as dimensionality-reductions to
different numbers of principal components.

 Several of the classifiers achieved high levels of performance,
reaching MCC scores of \textasciitilde{}0.6, F1 scores of \textasciitilde{}0.8,
and iAUC scores of \textasciitilde{}0.86. Bigrams in combination with
unigrams tended to perform better than unigrams alone, and the combination
of document-frequency and length normalization also tended to have
a slight positive effect on performance. This effect may have been
more pronounced if we had used count (instead of occurrence) matrices,
in which document vector magnitudes are more variable. In addition,
we also implemented PCA-based dimensionality reduction. Its effect
on performance was mild for most classifiers, except for linear discriminant
analysis (LDA). We observed dismal LDA performance with no dimensionality
reduction, and high performance when data matrices were projected
onto their first 800-1000 principal components. This is consistent
with the well-known weakness of LDA in high-dimensional classification
contexts.

 Both relevant and irrelevant training sets came from the field of
pharmacokinetics and, for this reason, shared very similar feature
statistics. This makes distinguishing between them quite a difficult
text classification problem -- though also a more practically relevant
one (such as in a situation where a researcher needs to automatically
label a pre-filtered a list of potentially relevant documents). It
may also explain why the NER features did not make a positive impact
on classification performance: the documents in both classes would
be expected to have similar counts of drug names, proteins, and other
named entities, and so these counts would not help class separation.
It is possible, of course, that the use of NER more finely tuned to
DDI, relation extraction, or some other more sophisticated feature-generation
technique could improve performance.

To conclude,  the best performing classifiers and feature-transforms
led to similar upper limits of performance, suggesting a fundamental
limit on the amount of statistical signal present in the labels and
feature distributions of the corpus. However, to achieve near-optimal
generalization performance, selecting the proper combination of classifier,
feature transforms, and dimensionality-reduction is necessary. When
working with classifiers that contain cross-validated parameters,
this can be done through the use of nested cross-validation.

Using such procedures, we show that, under realistic classification
scenarios, automatic BLM techniques can effectively identify reports
of DDIs backed by pharmacokinetic evidence in  PubMed abstracts.
These reports can be essential in identifying causal mechanics of
putative DDIs, and can serve as input for further \emph{in vitro}
pharmacological and \emph{in populo} pharmaco-epidemiological investigation.
Thus, our work provides an essential step in enabling further development
of interdisciplinary translational research in DDI.

\section{Acknowledgments}

This work was supported by the \emph{Indiana University Collaborative
Research Grant }``Drug-Drug Interaction Prediction from Large-scale
Mining of Literature and Patient Records.''

\bibliographystyle{ws-procs11x85}
\bibliography{writeupshort}

\end{document}